\documentclass[a4paper, 10pt, conference]{ieeeconf} 

\IEEEoverridecommandlockouts                              

\title{\LARGE \bf
Stinger Robot: A Self-Bracing Robotic Platform for Autonomous Drilling in Confined Underground Environments
}

\author{Haotian Liu, Laura Santas Moreu, Tobias Stenbock Andersen, Victor Vigars Puche and Matteo Fumagalli
\thanks{The authors are with the Department of Electrical and Photonics Engineering,
        DTU, Elektrovej, Building 326, 2800 Kgs. Lyngby, Denmark
        {\tt\small haliu@dtu.dk lsamo@dtu.dk tstan@dtu.dk vvipu@dtu.dk mafum@dtu.dk}}%
\thanks{This work has been supported by the European Union’s Horizon Europe Research and Innovation Programme under the Grant Agreement No.101138451}%
\thanks{The DTU Autonomous Systems Test Arena (ASTA) has provided hardware and testing facilities for the project.}
}

\usepackage{graphicx}
\usepackage{subcaption}
\usepackage{hyperref}
 
\begin{document}

\maketitle
\thispagestyle{empty}
\pagestyle{empty}

\begin{abstract}
 The increasing demand for critical raw materials has revitalized interest in abandoned underground mines, which pose extreme challenges for conventional drilling machinery due to confined, unstructured, and infrastructure-less environments. This paper presents the Stinger Robot, a novel compact robotic platform specifically designed for autonomous high-force drilling in such settings.
The robot features a mechanically self-locking tri-leg bracing mechanism that enables stable anchoring to irregular tunnel surfaces. A key innovation lies in its force-aware, closed-loop control strategy, which enables force interaction with unstructured environments during bracing and drilling. Implemented as a finite-state machine in ROS 2, the control policy dynamically adapts leg deployment based on real-time contact feedback and load thresholds, ensuring stability without external supports.
We demonstrate, through simulation and preliminary hardware tests, that the Stinger Robot can autonomously stabilize and drill in conditions previously inaccessible to nowadays mining machines. This work constitutes the first validated robotic architecture to integrate distributed force-bracing and autonomous drilling in underground environments, laying the groundwork for future collaborative mining operations using modular robot systems.
\end{abstract}




\section{Introduction}








The EU’s transition to a green and digital economy has sharply increased demand for critical raw materials (CRMs) like lithium, cobalt, and rare earth elements \cite{CRM, public_mines}. These materials are vital for technologies such as electric vehicles and wind turbines \cite{CRM_demand, CRM_battery}, yet their supply chains are highly import dependent \cite{CRM_nonEU} and vulnerable to geopolitical risks \cite{CRM}. To improve resilience, the EU is promoting recycling \cite{CRM_recycling}, substitution, and the reactivation of domestic sources, including abandoned underground mines. However, reopening these sites poses significant operational and safety challenges that require innovative technological solutions.

Deep and abandoned underground mines are highly complex environments, often unsafe for human access, characterized by narrow, irregular tunnels, unstable terrain, poor visibility, and complete lack of infrastructure. These conditions render traditional mining equipment, which relies on flat surfaces, spacious tunnels, and heavy machinery for stability, ineffective or even inoperable. Machines like face drill rigs and bolting platforms from companies such as \texttt{Epiroc} \footnote{See \href{https://www.epiroc.com/en-cz/products/drill-rigs}{https://www.epiroc.com/en-cz/products/drill-rigs}} require large, open working areas and deploy heavy hydraulic stabilizers to absorb the reaction forces generated during drilling. These designs are poorly suited for collapsed or constricted tunnels

Efforts to miniaturize mining machinery for operation in confined and irregular environments have led to the development of platforms such as the RM1 concept, introduced within the \textit{ROBOMINERS} project \cite{ROBOMINERS}. While significantly smaller than conventional drill rigs, RM1 remains a monolithic system with limited mobility in unstructured terrain and is primarily constrained to drilling in the forward direction. Alternative approaches, such as the larva-inspired drilling robot proposed by Liao et al. \cite{larva_drilling_robot}, exhibit promising locomotion capabilities in underground settings but are restricted to soft substrates and lack the structural robustness required for hard-rock drilling.
Outside the mining domain, some robotic systems have attempted to achieve mechanical stability by generating bracing forces through interaction with the environment. For example, Boomeri et al.\cite{similar_three_leg_robot} and Fu et al.\cite{tree_climbing_robot} developed climbing robots capable of anchoring to poles or trees, while Tao et al.\cite{strong_climbing_robot} introduced a vacuum-based climbing robot able to support high external loads on smooth vertical surfaces. However, these designs are either too environment-specific or unable to cope with the surface irregularities and force levels typical of deep underground operations.
While existing robotic platforms for exploration or climbing provide useful mobility, they generally lack the structural rigidity or adaptive bracing needed for high-force tasks like drilling. Many are designed for smooth surfaces or low-resistance environments and cannot compensate for the reaction forces or spatial constraints inherent to underground mining. As a result, a clear technological gap remains: the absence of a compact, mobile robotic system capable of autonomously stabilizing itself and executing precision drilling operations under extreme subterranean conditions.
In our original work \cite{next_generation_mining}, we proposed an innovative approach that leverages multi-robot technologies to break down large rigid machines into smaller, task-specific robotic elements that can operate collaboratively. Within this paradigm, this paper presents the \textit{Stinger Robot}—a compact, modular robotic platform that serves as the foundation for assemble drilling rigs and execute drilling operations. \\
This paper contributes the first validated architecture for a compact, self-stabilizing drilling robot that combines mechanically self-locking tri-leg bracing with a closed-loop, force-aware control policy tailored to unstructured underground environments. Unlike prior climbing or drilling platforms, the Stinger Robot adapts dynamically to uneven, infrastructure-less terrain using compliant contact feedback, enabling robust anchoring and precision drilling without human intervention or external support structures.\\
The contributions of the paper are:
\begin{itemize}
    \item A mechanically self-locking tri-leg configuration optimized for high-force axial resistance in confined geometries.
    \item A force-threshold-based bracing control strategy, integrated via a four-phase ROS 2 control pipeline, enabling autonomous anchoring under variable surface conditions.
    \item The first prototype validation of force-distributed, contact-adaptive self-bracing in a compact robotic platform suitable for hard-rock subterranean operations.
\end{itemize}
The paper furthermore presents a ROS 2-based modular simulation and control stack for distributed, cooperative underground drilling missions.


    







\section{Stinger Robot Design}

\subsection{Concept of Operation and Stinger Robot Function}
To overcome the challenges posed by underground mining environments, we adopt a modular, multi-robotic approach, first introduced in \cite{next_generation_mining}, in which task-specific robots operate collaboratively to achieve autonomous drilling in  extreme mining settings. This modularity allows each robot to remain compact and specialized, while collectively addressing the demands of a complex mission that would otherwise exceed the capabilities of any single platform.

In this framework, the drilling mission begins with autonomous robots exploring previously unmapped tunnel sections and building a 3D representation of the environment. These robots analyze the terrain to identify promising drilling locations based on spatial and geological criteria. Once a suitable site is selected, a dedicated deployment robot navigates to the area and uses its onboard perception and manipulation capabilities to analyze the local geometry and position the Stinger robot precisely at the drilling location. The Stinger then anchors itself to the tunnel surface using its three-legged bracing mechanism, ensuring mechanical stability in preparation for drilling. The stinger robot braced within the walls of the mine functions as a reliable and stable base for mounting a drill rig and subsequently perform the drilling operation. During this process, a support robot may assist by delivering essential supplies such as power units, replacement tools, or drilling fluids. After completing the task, the deployment robot can reposition the Stinger to repeat the operation at a new site.

In this mission structure, the Stinger robot plays a central role as the only agent capable of physically providing a mechanically robust basis for mounting a drill rig. The functionalities that the stinger robot design should feature are: (1) to robustly attach to the mine walls; (2) to ensure vibrations are transfered to ground, without overloading joints and gearboxes; (3) allow precise placement of an extention drilling unit which will complete the drilling operation. Its integration into this modular and collaborative pipeline enables efficient, safe, and autonomous drilling in hazardous environments that would be inaccessible or too dangerous for human workers, and impossible to reach for nowadays drilling machines, due to their sizes.

The remaining of the section shows how these functionalities are achieved by means of analysis, simulation, physical prototyping and partial deployment.

\subsection{Kinematic and Workspace Analysis}
To guarantee stable and robust drilling, a three legged system is chosen, intended to form a closed loop kinematic chain when all legs are in contact with the unvironment. To ensure that the robot can be deployed and brace on the mine's wall, all thee legs comprise a linear actuator that allow the legs to extend and reach the mine walls. In order to ensure that the robot can be deployed in potentially any confined concave mine section, two of the three legs will comprise a rotational actuator that allows rotating the legs w.r.t. themain body, thus enabling a wider set of configurations that can be selected for deployment and for optimizing the bracing configuration. Therefore, the Stinger Robot comprises  five actuated degrees of freedom. When the anchor points are in contact with the environment, the robot forms a closed kinematic chain. Given the robot's topology, its workspace is constrained by the minimum and maximum extension lengths and rotation angles of each stinger. Since the central stinger is rigidly attached to the main body, the robot's orientation aligns with that of the central stinger. The overall reachable workspace is determined by the positions of the fixed anchor points. 

A global coordinate frame is established at the anchor point of the central stinger leg, which serves as the robot’s base reference. Each stinger’s configuration is parameterized by its extension length $l_i$ and rotation angle $\theta_i$, both measured relative to this central coordinate frame. Each stinger's extension length \( l_i \) is constrained within the limits:
$$
l_i \in [\,l_{\min},\, l_{\max}\,]
$$
where $l_{\min}=625mm$ and $l_{\max}=1125mm$. Its rotation angle \( \theta_i \) of each side stinger is bounded by:
$$
\theta_i \in [\,\theta_{\min},\, \theta_{\max}\,].
$$
where $\theta_{\min}=-90^\circ$ and $\theta_{\max}=90^\circ$. 

\begin{figure}[h]
  \centering
  \begin{subcaptionbox}{\label{fig:min_extension}}[0.3\linewidth]
    {\includegraphics[width=\linewidth]{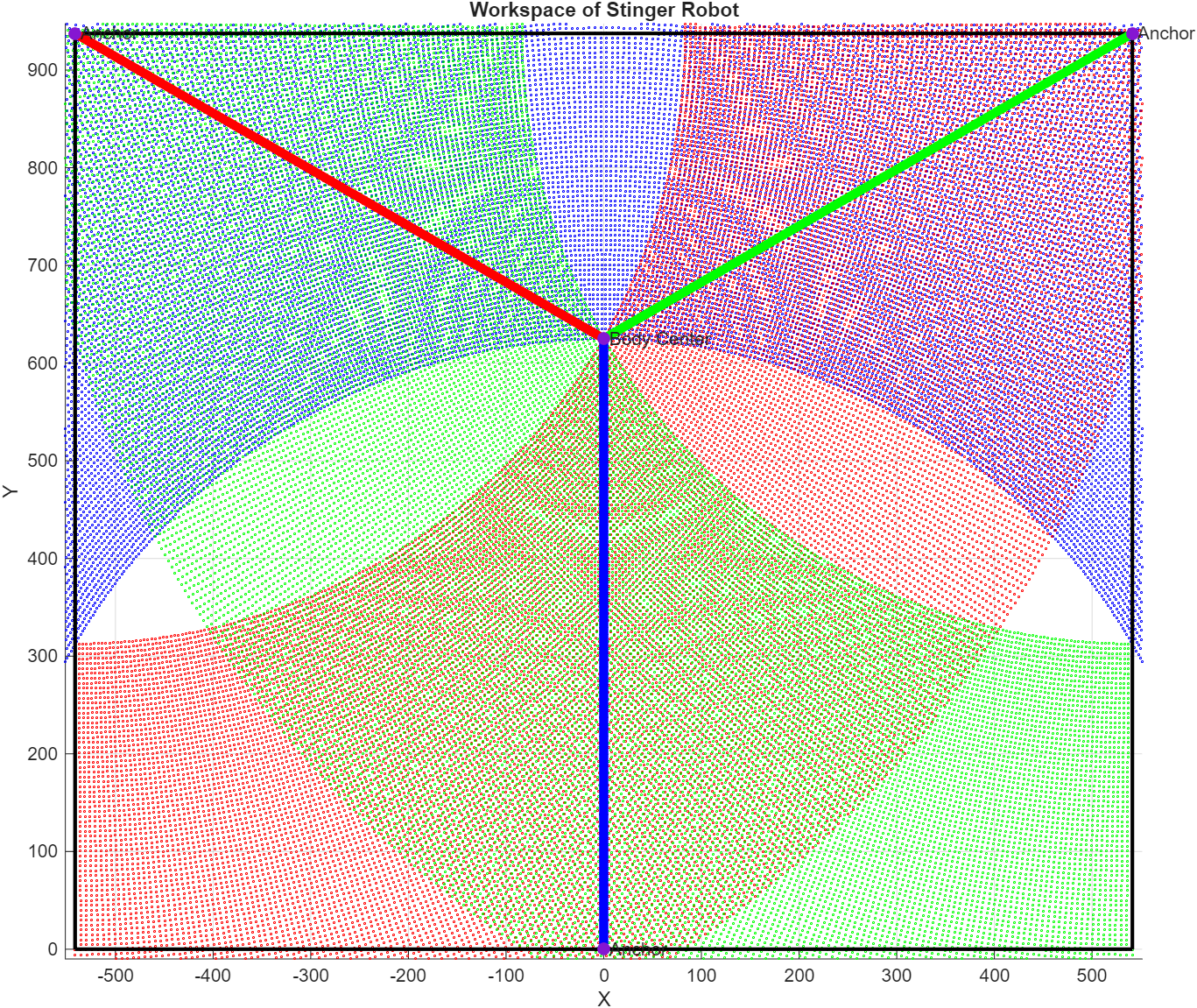}}
  \end{subcaptionbox}
  \hfill
  \begin{subcaptionbox}{\label{fig:max_extension}}[0.3\linewidth]
    {\includegraphics[width=\linewidth]{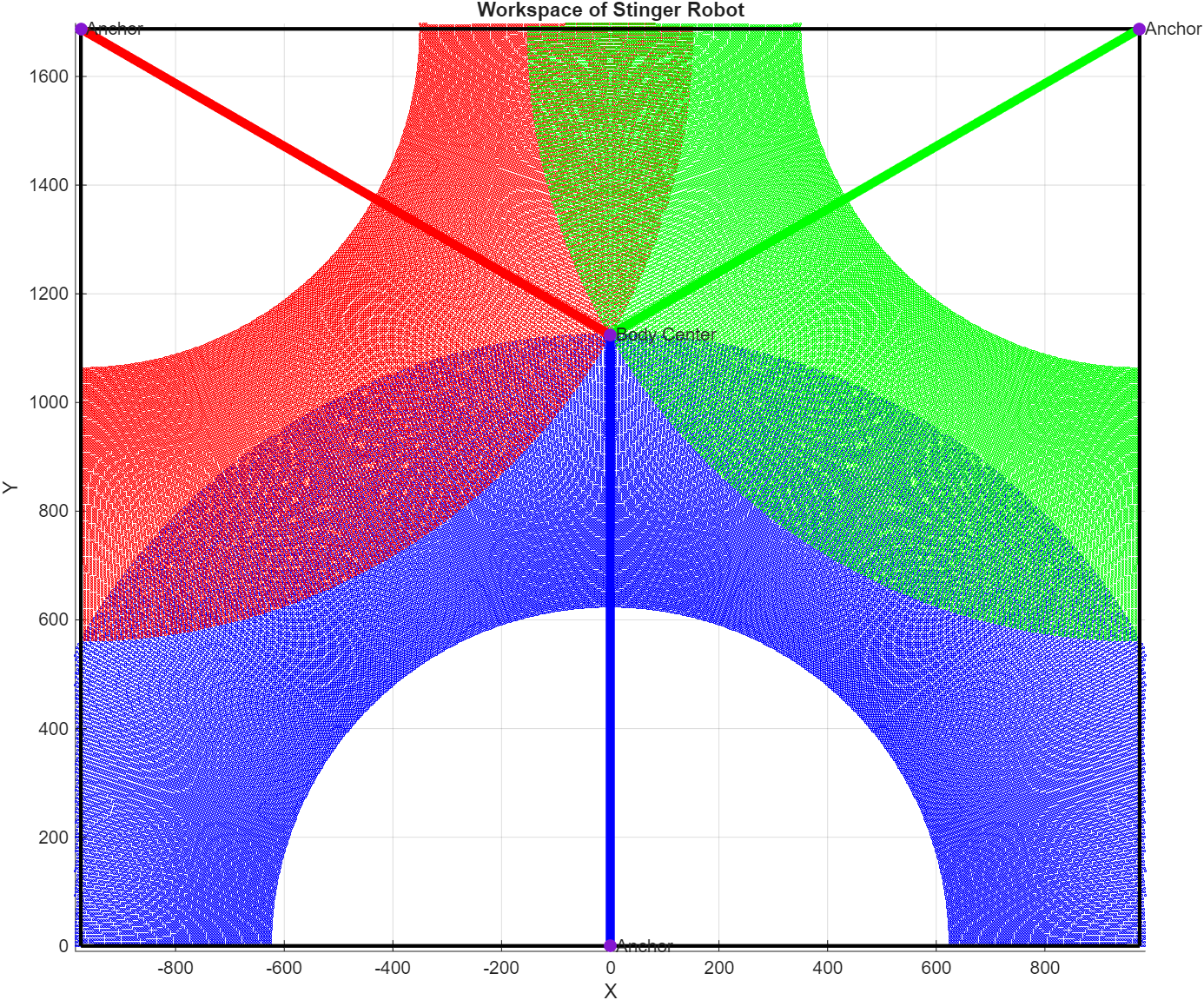}}
  \end{subcaptionbox}
  \hfill
  \begin{subcaptionbox}{\label{fig:extension_1}}[0.3\linewidth]
    {\includegraphics[width=\linewidth]{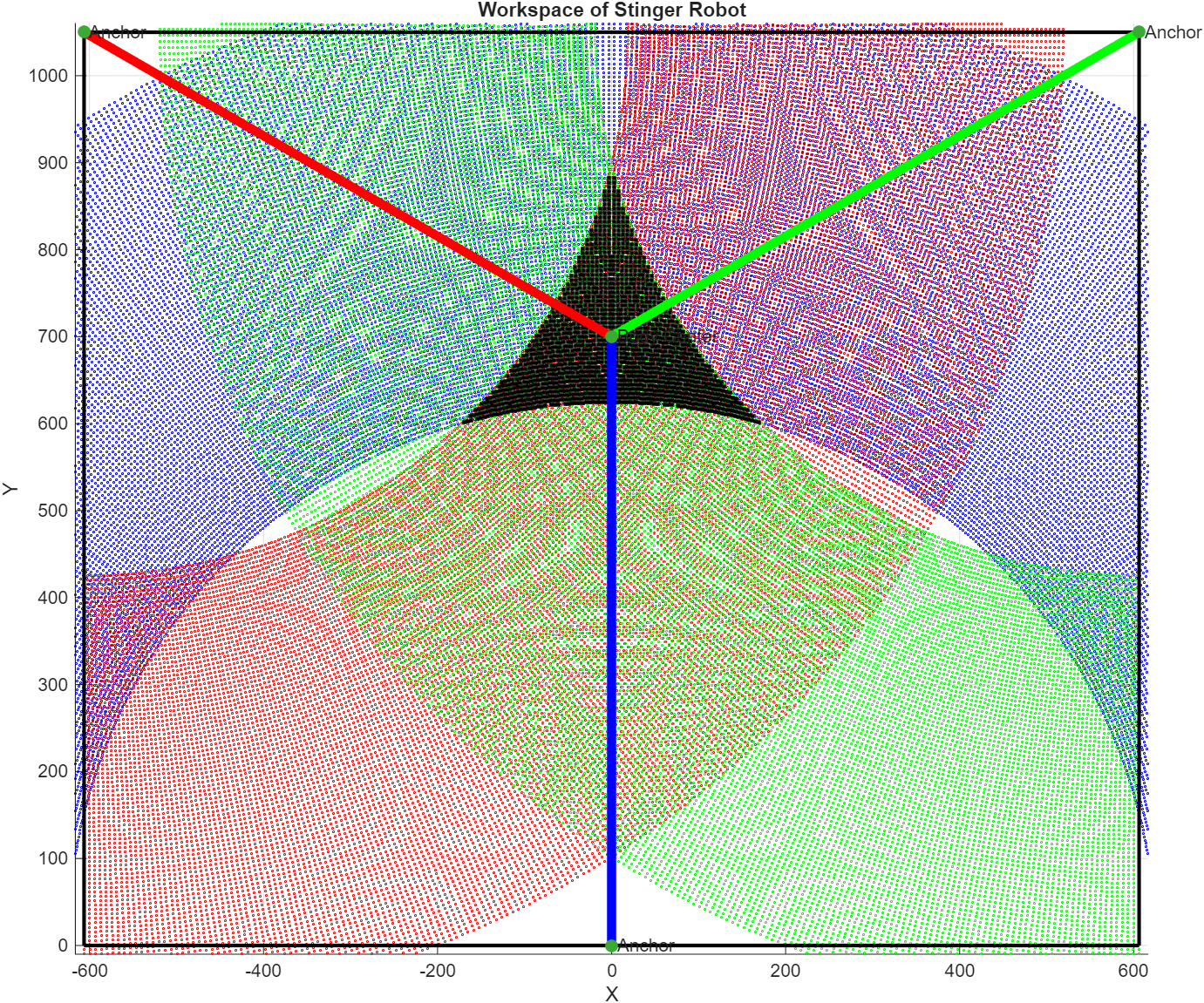}}
  \end{subcaptionbox}
  \begin{subcaptionbox}{\label{fig:extension_2}}[0.3\linewidth]
    {\includegraphics[width=\linewidth]{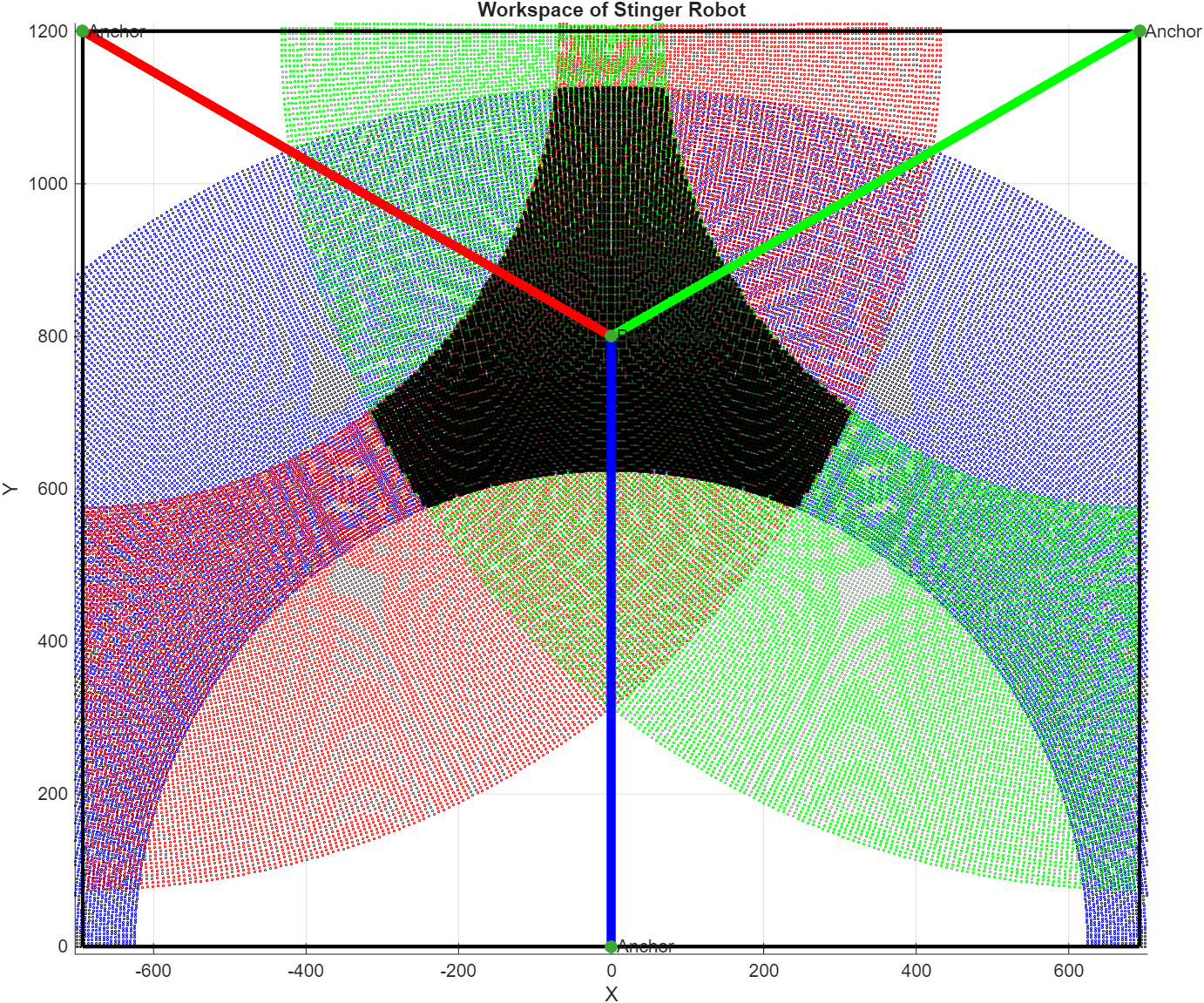}}
  \end{subcaptionbox}
  \begin{subcaptionbox}{\label{fig:extension_3}}[0.3\linewidth]
    {\includegraphics[width=\linewidth]{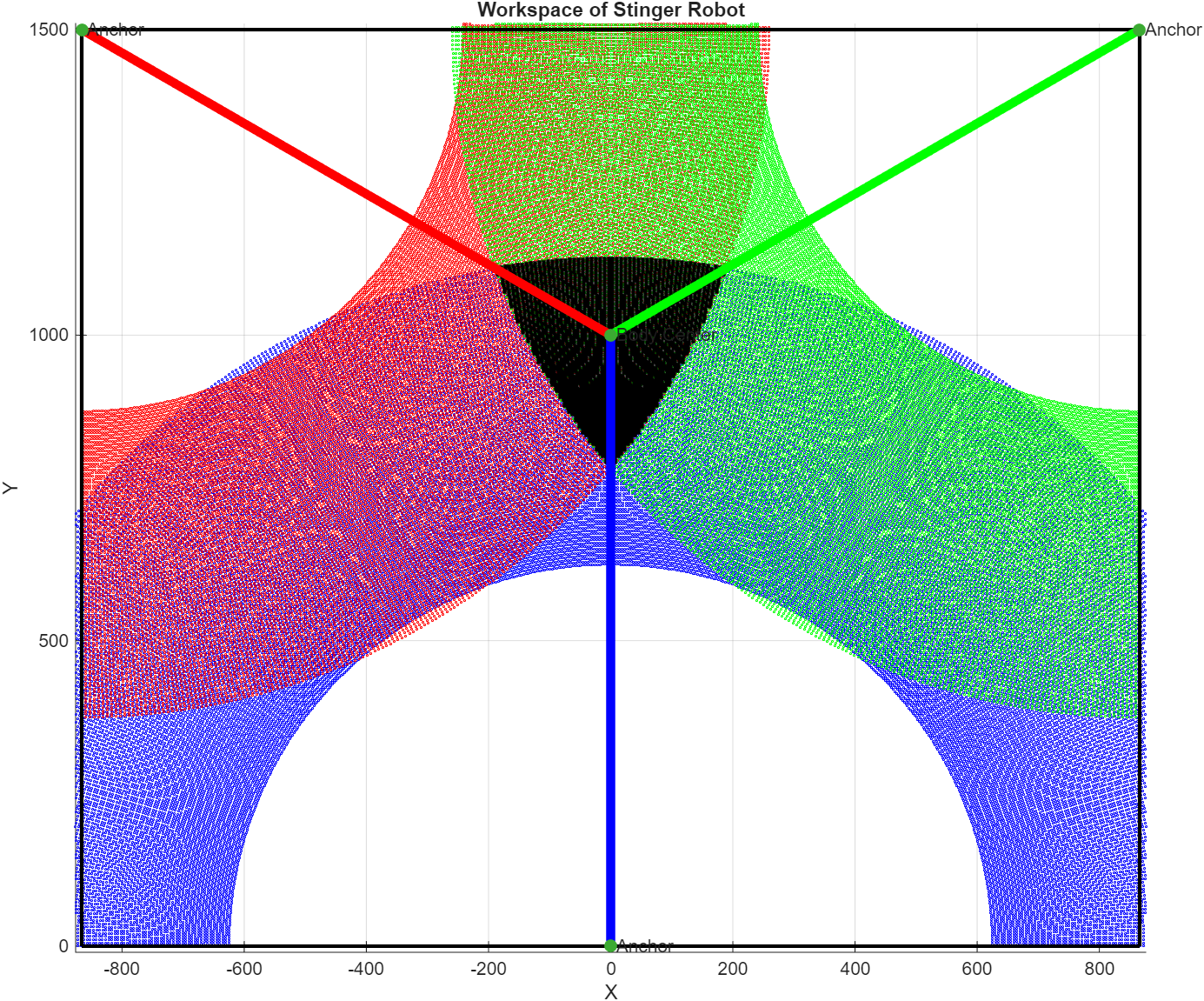}}
  \end{subcaptionbox}
  \begin{subcaptionbox}{\label{fig:extension_vertical}}[0.3\linewidth]
    {\includegraphics[width=\linewidth]{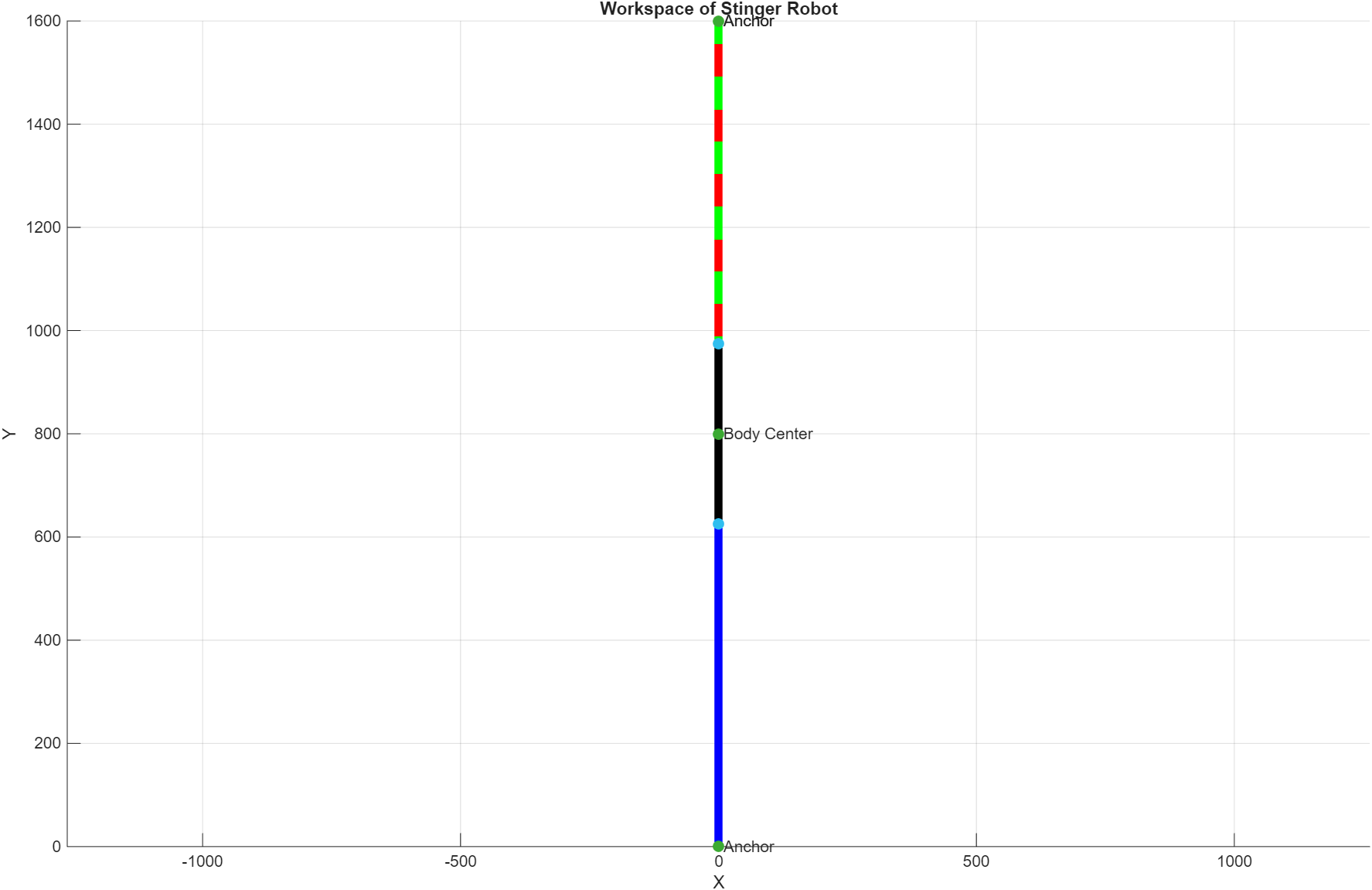}}
  \end{subcaptionbox}
  \caption{Workspace configurations of the Stinger Robot under various stinger extension and orientation scenarios. (a) Stingers at minimum extension.(b) Stingers at maximum extension.(c–e) Stingers extended to intermediate positions between minimum and maximum.(f) Side stingers rotated by 180$^\circ$. (a) and (b) show the robot with minimum and maximum stinger extensions, respectively, resulting in a single reachable point in each case. (c), (d), and (e) illustrate intermediate extensions that generate a continuous reachable workspace area. (f) depicts a special case where both side stingers are rotated by 180°, producing a linear workspace along the central stinger's longitudinal axis. }
  \label{fig:workspace}
\end{figure}

Figure~\ref{fig:workspace} illustrates various workspace configurations of the Stinger Robot under different deployment scenarios. In the kinematic analysis, the Stinger Robot is simplified for clarity. The blue line represents the central stinger, while the red and green lines correspond to the left and right side stingers, respectively. Each colored ring denotes the reachable workspace of the associated stinger. The overlapping region, shown in black, indicates the effective workspace of the central body. 


It is noticeable that the achievable workspace—and thus the robot’s ability to reposition its central body to align with the drilling task—is highly dependent on both the deployment configuration and the size and morphology of the mine section in which it operates. As illustrated in Figure \ref{fig:workspace}, if the mine cross-section is too large, the robot’s stingers must fully extend to achieve a stable configuration, resulting in a drastically reduced workspace, effectively limited to a single point. Conversely, if the mine section is too narrow, the robot's central body becomes constrained, unable to reposition due to insufficient clearance for leg articulation. In intermediate scenarios, where the mine section dimensions fall within a usable range, the circular workspaces of each stinger overlap, forming a usable region that enables body repositioning. This effective workspace is thus a function of both the leg deployment configuration and the geometry of the surrounding environment.

\subsection{Simuation}\label{simulation}


To test the robot deployment within a robot mission, as well as to analyze the stinger's force capability to drill (see Section \ref{sec:experiments}, a physics-based simulation of the Stinger Robot was developed in ROS 2 and Ignition Gazebo, using a modular robot model defined in URDF. The robot consists of a compact central body that supports the mounting of three deployable stinger legs and a front-facing dummy drill unit. The two side stingers are connected via revolute joints for angular positioning, enabling stable wall contact in varying tunnel geometries.
Each stinger includes a prismatic joint for extension, controlled via velocity commands, while revolute joints use position control, all implemented with ROS 2 Control \footnote{See \href{https://github.com/ros-controls/ros2\_control}{https://github.com/ros-controls/ros2\_control}}. The drill unit features a revolute joint for alignment and a prismatic joint to simulate contact and penetration during drilling. To analyze force distribution, force sensors are placed at the tip of each leg and at the drill bit. This setup enables testing of bracing stability, force distribution, and drilling interactions in confined environments, laying the groundwork for subsequent experimental validation.

\subsection{Mechatronic Design}

The central body of the robot is designed to interface with a compact, mechanically self-locking bracing system composed of three deployable legs. These "stingers" provide anchoring capability by pressing against the surrounding rock, enabling stable operation during high-force drilling. Drawing from principles found in large-scale industrial drilling anchors, the system balances structural rigidity with spatial efficiency, making it well-suited for deployment in confined underground environments.

\begin{figure}[h]
  \centering
  \begin{subcaptionbox}{Front View of the Stinger Robot\label{fig:front_view_stinger}}[0.45\linewidth]
    {\includegraphics[width=\linewidth]{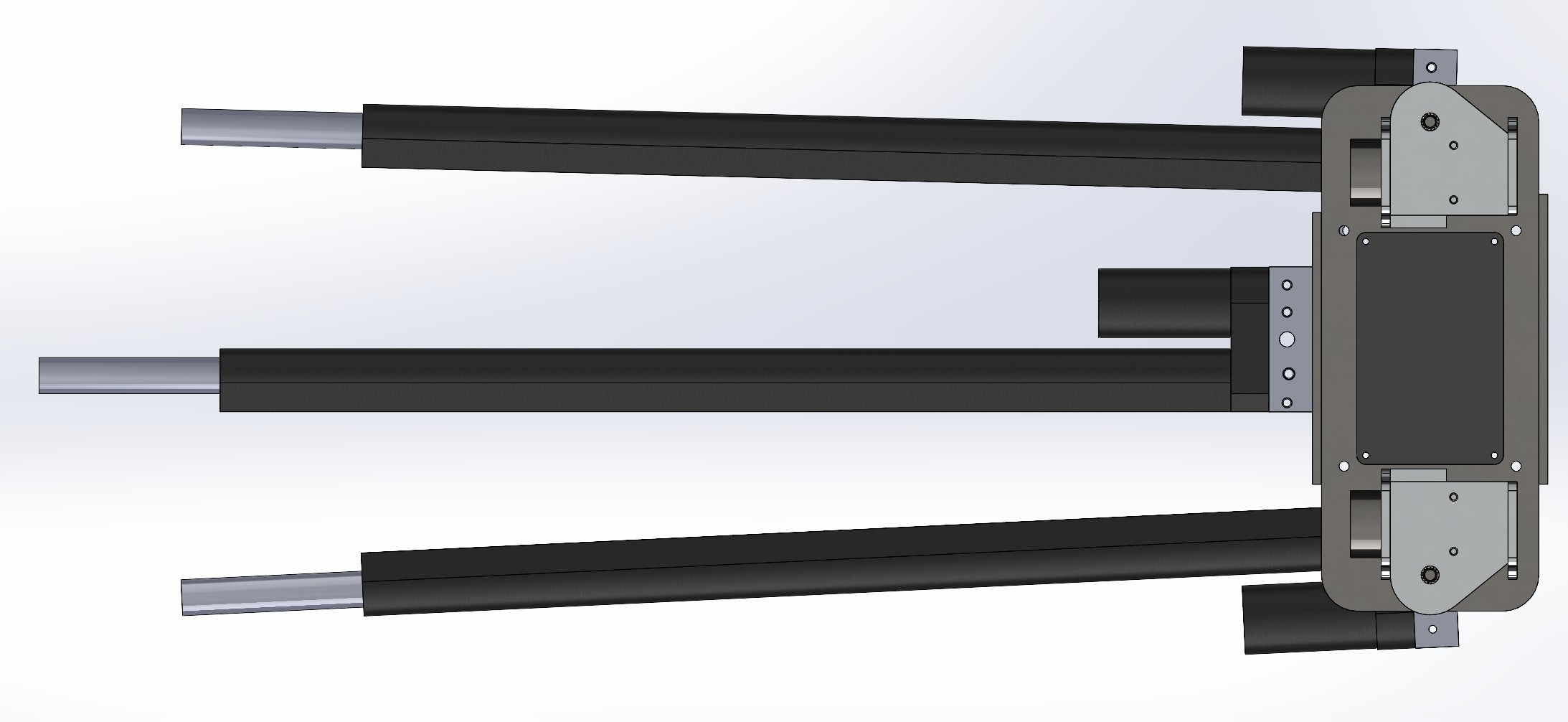}}
  \end{subcaptionbox}
  \hfill
  \begin{subcaptionbox}{Top View of the Stinger Robot\label{fig:top_view_stinger}}[0.45\linewidth]
    {\includegraphics[width=\linewidth]{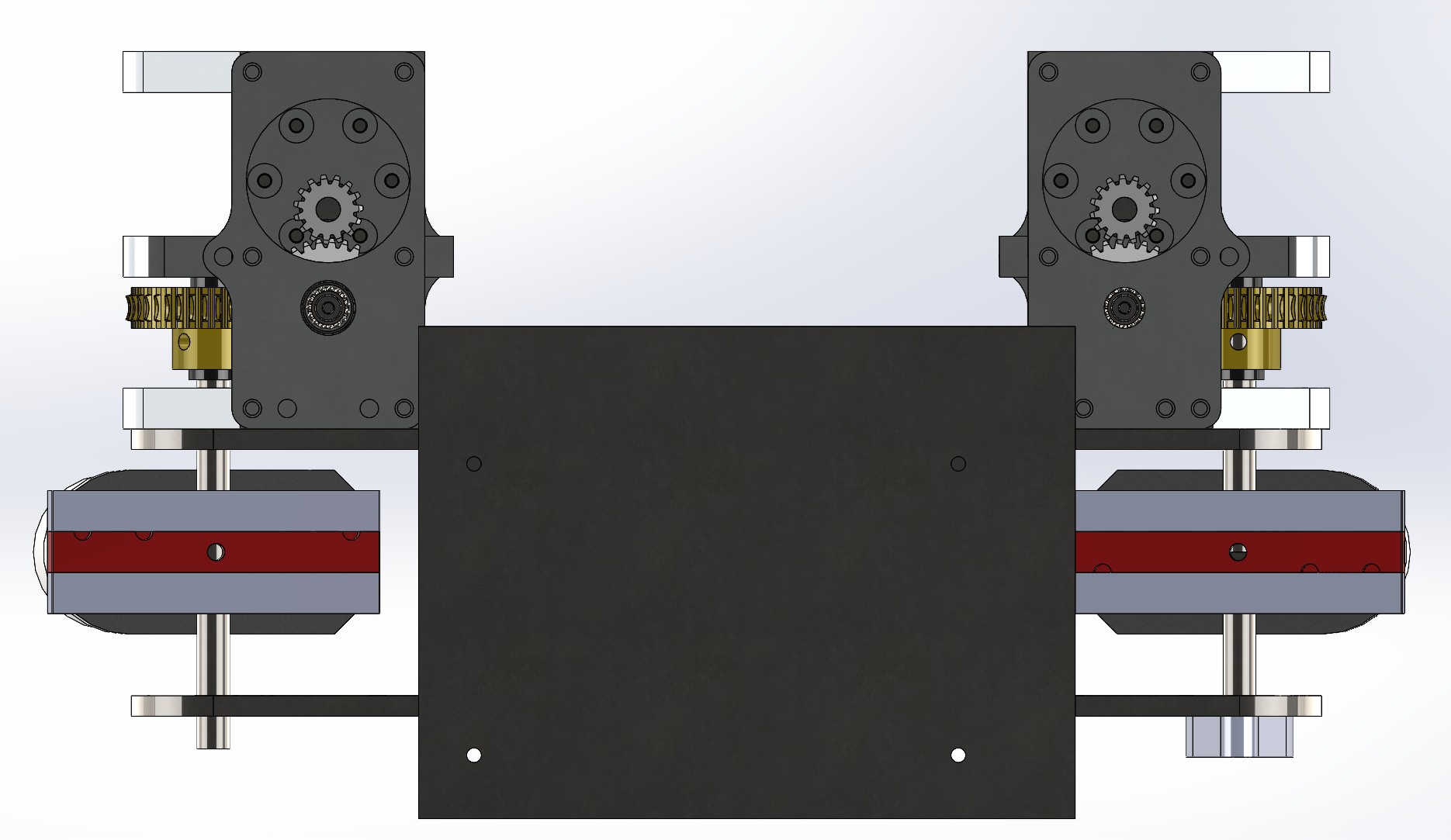}}
  \end{subcaptionbox}
  \hfill
  \begin{subcaptionbox}{Side View of the Stinger Robot\label{fig:side_view_stinger}}[0.45\linewidth]
    {\includegraphics[width=\linewidth]{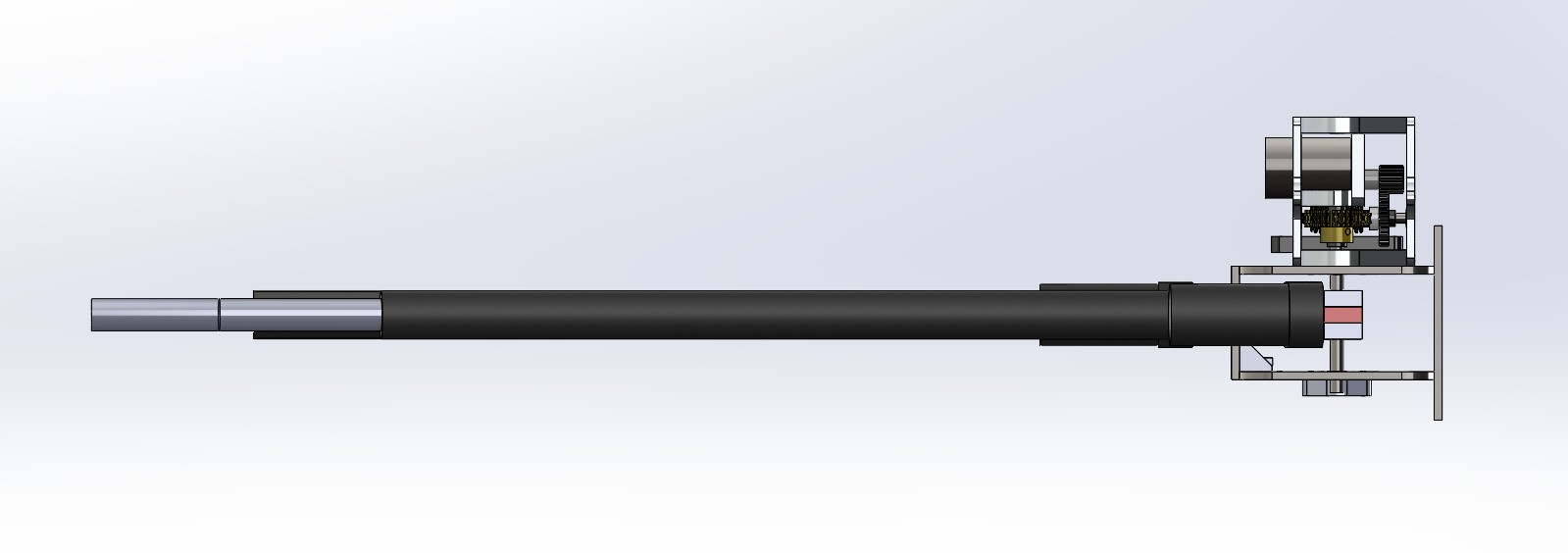}}
  \end{subcaptionbox}
  \caption{CAD views of the Stinger Robot.
(a) Front view, (b) Top view, and (c) Side view of the robot showing the three-legged self-bracing mechanism. The central body houses the transmissions for reconfiguring the legs' orientation and electronics. The two side stingers are mounted via rotary joints for angular deployment. The configuration enables stable anchoring against tunnel walls, allowing the robot to maintain its position during drilling operations in confined underground environments. The body will host the mechatronic interfaces for adding the drilling unit.}
  \label{fig:stinger_CAD}
\end{figure}

The Stinger robot incorporates five degrees of freedom, distributed across three deployable bracing legs. Each stinger extends via a linear actuator; the two lateral stingers are mounted on rotary joints, allowing up to 180° of motion, while the central leg is fixed to provide direct axial support. High-ratio gearboxes (80:1) drive the rotational actuation of the side legs, enabling controlled deployment within confined geometries. This configuration supports flexible yet stable anchoring across a range of tunnel shapes. The complete CAD model and orthogonal views of the robot are presented in Figures \ref{fig:stinger_CAD}.

The central linear actuator is rigidly attached to the robot body and offers linear motion only. In contrast, the side actuators possess both linear motion and rotational capability. All linear motion is driven by VEVOR linear actuators. Each actuator has a total length of 625 mm, with a stroke length of 500 mm, and a maximum pushing force of 1000 N.


The rotary joints of the side stingers use a custom two-stage reduction system, composed of a 1:2 spur gear followed by a 1:40 worm gear, yielding an overall gear reduction of 80:1. This configuration provides high torque output and mechanical simplicity. Due to the large helix angle of the worm, the worm gear cannot backdrive the worm when under load, thus having an inherent self-locking capability, which enhances safety.

The worm gearbox is driven by a DFROBOT 12V metal DC geared motor, which has a stall torque of 1,765Nm. 


Experimentally, spur gear systems exhibit an efficiency of approximately 95–98\%, while worm gear systems typically range between 40\% and 90\% \cite{worm_gear_efficiency} due to their inherent sliding motion. Based on this, the estimated overall efficiency of the worm gearbox is between 38\% and 88.2\%.

According to equation \ref{eq: torque_cal}:
\begin{equation}
    T_{out} = T_{in} \times R \times \eta
    \label{eq: torque_cal}
\end{equation}
where $T_{out}$ is the output torque, $T_{in}$ is the input torque, $R$ is the gear ratio, and $\eta$ is the efficiency, the conservative estimate of the output torque is 53.656Nm.

\begin{figure}[h]
  \centering
  \begin{subcaptionbox}{CAD of the Gearbox\label{fig:gearbox}}[0.45\linewidth]
    {\includegraphics[width=\linewidth]{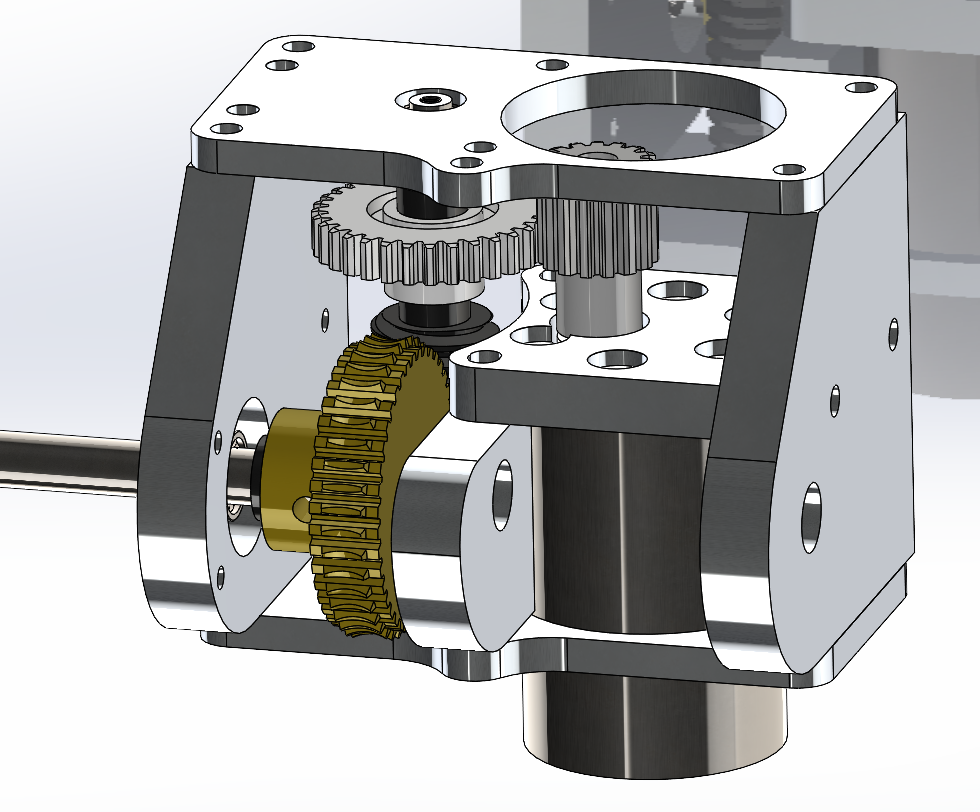}}
  \end{subcaptionbox}
  \hfill
  \begin{subcaptionbox}{Exploded View of the Gearbox\label{fig:gearbox_exploded}}[0.45\linewidth]
    {\includegraphics[width=\linewidth]{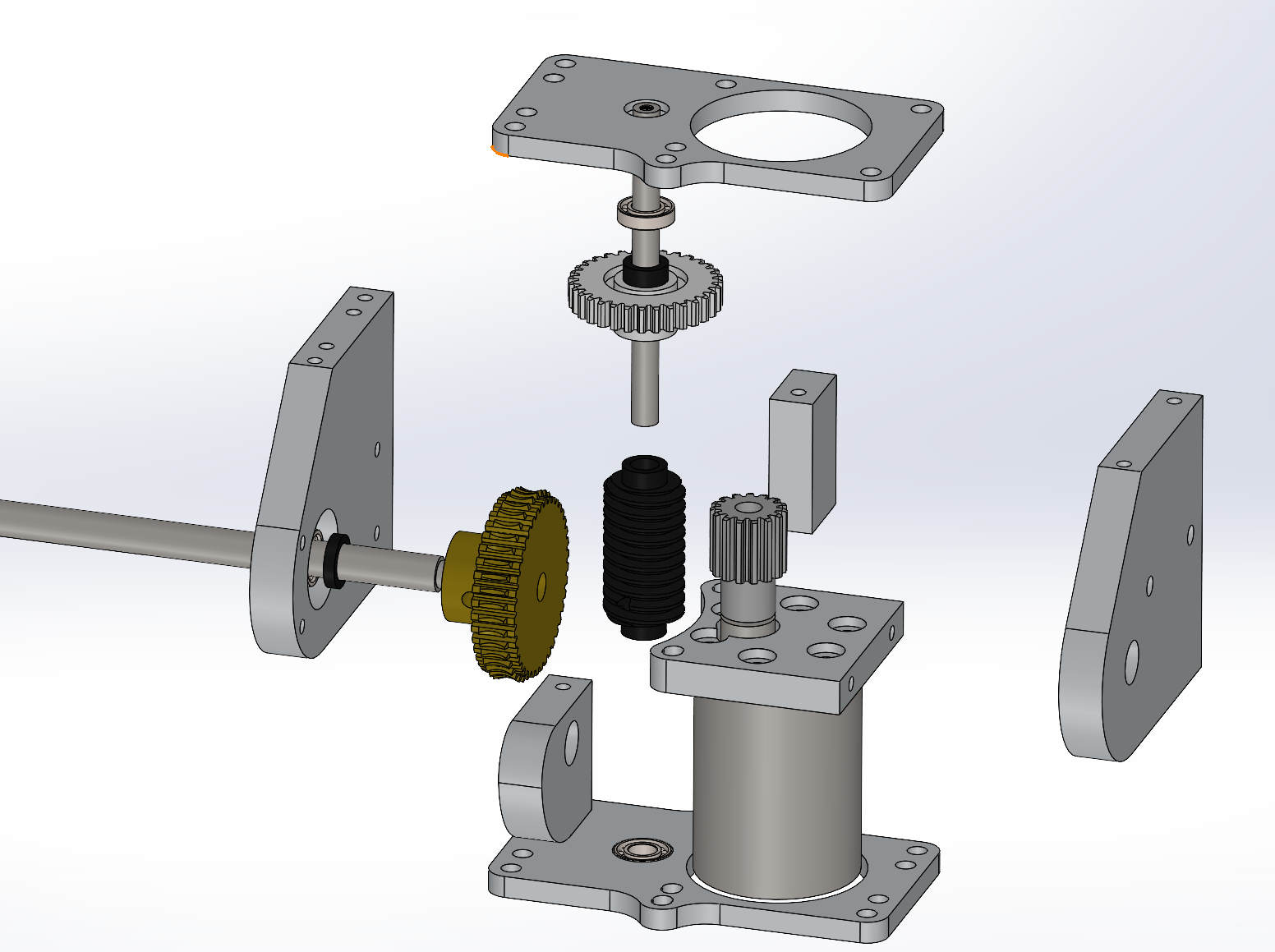}}
  \end{subcaptionbox}
 \begin{subcaptionbox}{\label{fig:gearbox_real}}[0.45\linewidth]
    {\includegraphics[width=\linewidth]{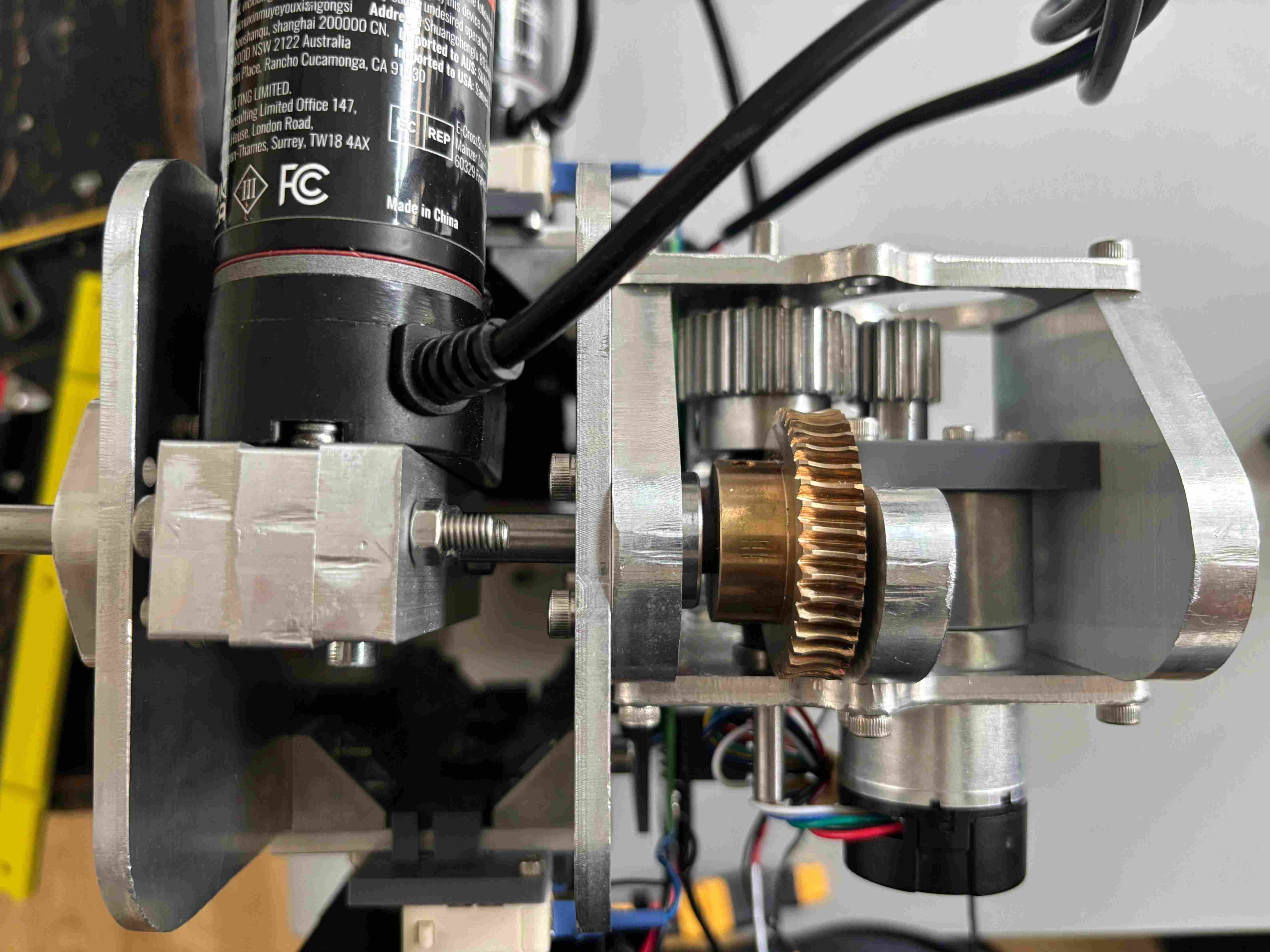}}
  \end{subcaptionbox}
  \hfill
  \begin{subcaptionbox}{\label{fig:gearbox_wheel}}[0.45\linewidth]
    {\includegraphics[width=\linewidth]{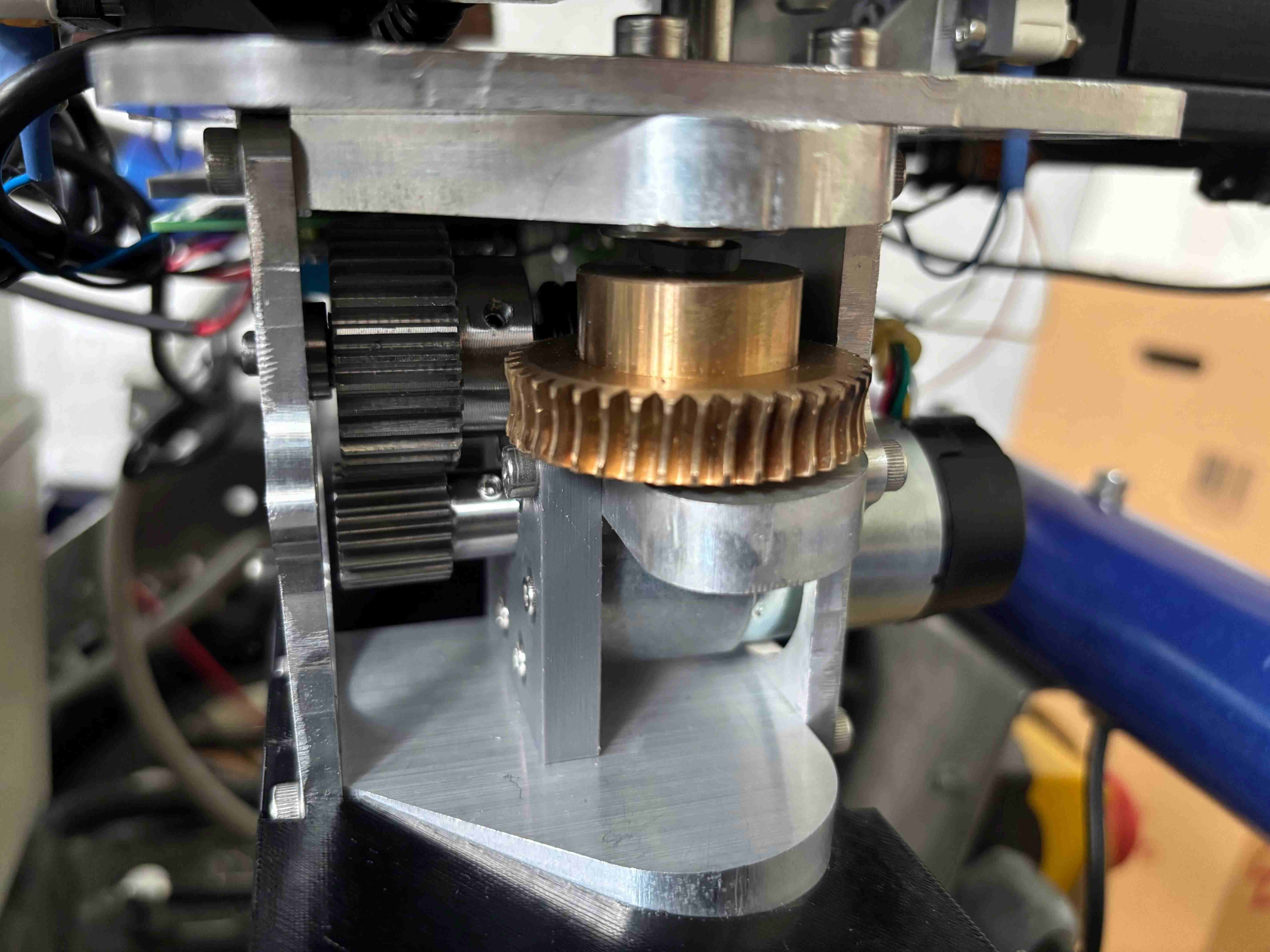}}
  \end{subcaptionbox}
  \caption{CAD model and exploded view of the worm gearbox used in the Stinger Robot.
(a) Assembled CAD model showing the compact aluminum housing; (b) Exploded view illustrating the internal components, including the spur gear pair and 1:40 worm gear stage. This two-stage reduction system provides high torque output and inherent self-locking capability, ensuring stable positioning of the rotary stinger joints during bracing and drilling operations; (c) and (d) Views of the prototyped joint.}
  \label{fig:gearbox_CAD_real}
\end{figure}

Figure. \ref{fig:gearbox_CAD_real} (a-b) shows the CAD design of the worm gearbox, while Figure \ref{fig:gearbox_CAD_real}(c-d) shows the manufactured gearbox. . 

\begin{figure*}[t]
    \centering
    \includegraphics[width=0.7\linewidth]{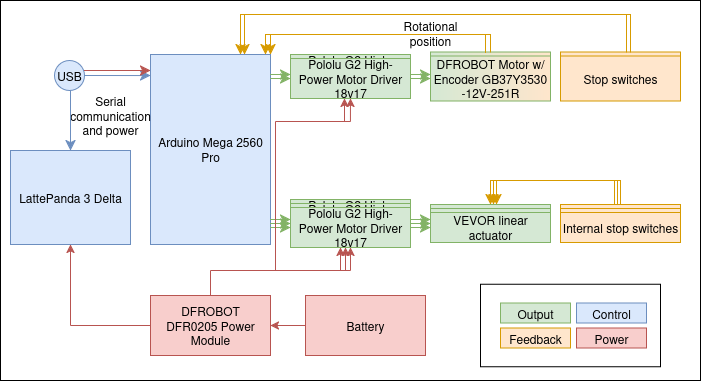}
    \caption{Overview of the Stinger Robot’s electronic architecture.
The system consists of a LattePanda 3 Delta for high-level ROS 2-based control and an Arduino Mega 2560 Pro for low-level actuator and sensor interfacing. Motor drivers control the linear and rotary actuators, while encoder feedback and limit switches enable precise joint positioning and safe operation. A single 12V power supply unit provides regulated power to all components.}
    \label{fig:electronics overview}
\end{figure*}

Motors are controlled following the system diagram of Figure \ref{fig:electronics overview} and interfaced to other mission agents through the ROS2\cite{ros2} middleware, which runs on the companion computer (a A LattePanda 3 Delta). 
The LattePanda is connected via a USB cable to an Arduino Mega 2560 Pro, and the two devices utilize serial communications. The Arduino handles the low-level control of the actuators and the communication with the sensors. 
Each actuator is driven by Arduino via a Pololu G2 High-Power Motor Driver 18v17. This allows for modulation of the input signals and, therefore, finer control of the actuators.
The two rotational joints mount motors that include encoders, which are read by the Arduino and, with the gearing of the rotational joint, give an accurate joint position estimate. A set of calibration switches has been included at the ends of the rotational stingers' operational limits. These safety switches connect back into the Arduino, and are used to block movements that would violate these limits. All electronic components are powered by a single DFR0205 Power Module calibrated at 12V. The power module performs voltage regulation and it ensures a steady power supply.

\section{Control Strategy}\label{control_strategy}
To enable autonomous deployment and drilling in unstructured underground environments, we define a four-phase control policy that governs the sequential behavior of the Stinger Robot. This control architecture integrates force feedback from embedded sensors at the tip of each stinger leg and the drill bit, enabling robust state transitions based on contact detection and load thresholds.

The full control strategy is implemented as a finite-state machine with four main operational phases:
\begin{enumerate}
    \item Opening Phase: The two side-mounted legs rotate outward via position-controlled revolute joints, forming a stable custom shape configuration (see Fig. X). This motion ensures the side stingers are optimally positioned to reach the surrounding tunnel surfaces.
    \item Initial Bracing Phase: Once oriented, all three prismatic legs begin to extend outward using velocity control. The control node continuously monitors real-time force feedback from each leg. When contact is detected (i.e., when measured force exceeds a predefined threshold), the corresponding leg halts. This behavior ensures light, distributed contact with the environment and marks the robot’s initial anchoring.
    \item Hard Bracing Phase: After contact is confirmed for all three legs, the robot enters a secondary, slow-extension mode where the prismatic actuators push further at reduced speed. A force-threshold feedback loop is applied: each leg continues extending until its contact force exceeds a second, higher limit and remains above it for a sustained duration. This ensures the robot achieves a stable, mechanically braced configuration capable of resisting external loads.
    \item Drilling Phase: Following stabilization, the robot aligns the drill with the wall via a revolute joint and extends it using a prismatic actuator. Once contact is detected at the drill bit, the system initiates rotation. Drilling proceeds until one of the stinger legs registers an excessive force beyond a safety limit, upon which the system halts to prevent structural overload.
\end{enumerate}

The entire control policy is implemented as a finite-state machine embedded within a ROS 2 node. State transitions are triggered by closed-loop feedback conditions based on joint positions and force thresholds, as shown in the experimental simulation plots in Section \ref{simulation_results}. This structure enables reactive, autonomous behavior without external supervision, ensuring that the robot adapts its deployment and drilling sequence to the surrounding geometry and contact dynamics in real time.

\section{Experiments}
\label{sec:experiments}
Two experiements were conducted to examine the capabilities of the Stinger Robot. A tension test to demonstrate how the robot is able to support itself via tension, and a step response test examining the accurateness of the stinger estimation through the encoder.


\subsection{Self-Bracing and Drilling Simulation Test}\label{simulation_results}
To evaluate the feasibility of the proposed control strategy, a complete simulation test was conducted using the setup explained in Section \ref{simulation}. The goal is to validate whether the Stinger Robot could autonomously execute the four-phase control policy introduced in Section \ref{control_strategy}: (i)~opening, (ii)~initial bracing, (iii)~hard bracing, and (iv)~drilling. The test replicates a realistic deployment scenario, where the robot is placed at a target location and autonomously transitions through each phase based on position and force feedback.

\begin{figure}[!h]
    \centering
    \includegraphics[width=0.45\textwidth]{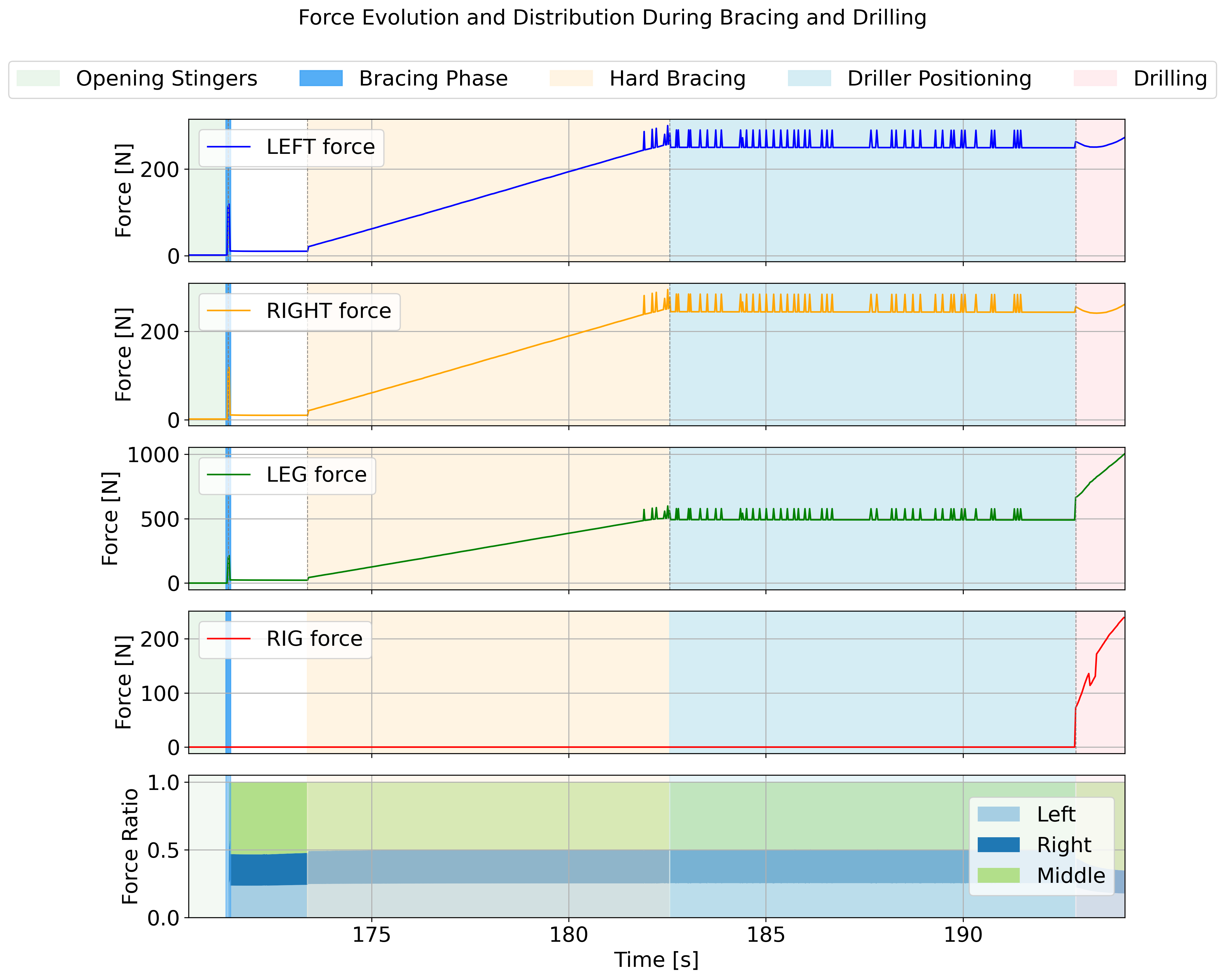} 
    \caption{Time-series plots showing the force magnitude at each of the Stinger Robot’s three anchoring legs and the drill bit, along with the force ratio distribution across the legs. Colored background regions highlight key mission phases.}
    \label{fig:sim_forces}
\end{figure}

Figure~\ref{fig:sim_forces} shows stacked force plots for each leg and the drill bit throughout the mission, along with a force ratio plot illustrating how the load is distributed among the three legs over time. Background colors annotate the mission phases, providing visual insight into the robot’s progressive stabilization and force distribution.

After the initial bracing phase, the robot achieves stable anchoring with relatively low contact forces: approximately 10\,N on the left and right legs, and 20\,N on the central leg. During this phase, the force distribution is approximately 50\% through the central leg and 25\% through each side leg, reflecting a symmetric bracing configuration centered along the drilling axis. In the hard bracing phase, these forces increase to approximately 140\,N on the side legs and 280\,N on the central leg, maintaining the same 50/25/25 load split while increasing total contact pressure to enhance stability. Once drilling begins, the contact forces rise significantly to absorb the reaction loads: the central leg bears the highest load at 1000\,N, while the side legs support 273\,N and 261\,N, respectively. This shift illustrates how the central leg increasingly absorbs the reactive force from the drill---mounted directly along its axis---while the contribution of the side legs slightly decreases. The drill bit also registers a surface contact force of 239\,N during this phase.

These results validate the effectiveness of the implemented control strategy in achieving autonomous self-bracing and stable drilling under force-feedback constraints. They demonstrate that the robot is capable of transitioning between control states, responding to contact events, and distributing structural loads as intended in unstructured, confined scenarios.

\subsection{Tension test}
\begin{figure}[!h]
    \centering
    \includegraphics[width=0.45\textwidth]{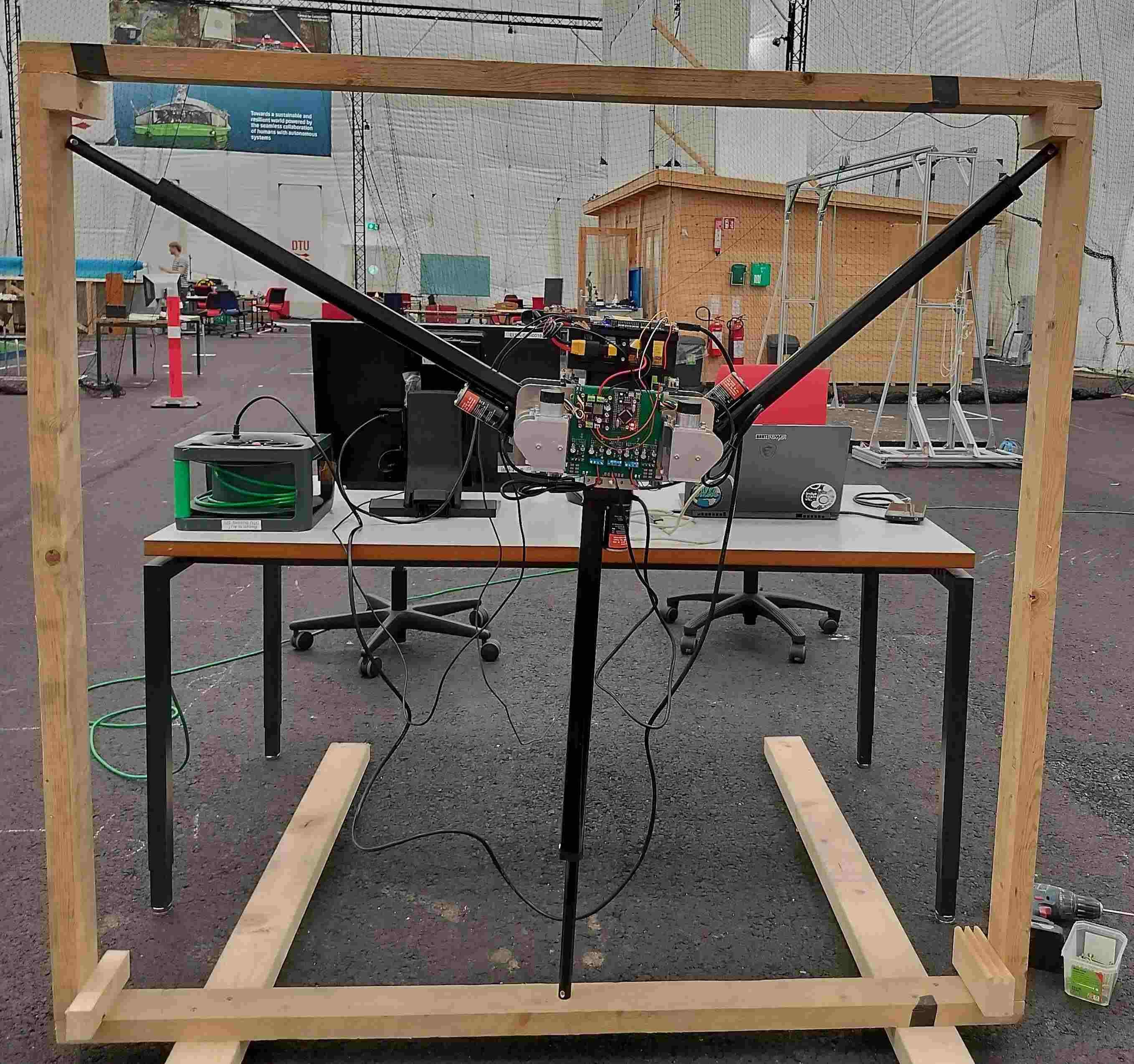} 
    \caption{The Stinger Robot deployed inside a test frame, demonstrating its bracing configuration in preparation for a simulated drilling task.}
    \label{fig:stinger_frame}
\end{figure}

A wooden frame was constructed to validate the Stinger Robot design. The Stinger Robot was then held in place while it was changed into a Y-shaped shape. The individual stingers were then slowly extended until they reached the frame. Once the stinger made contact with all three points, the linear actuators were stopped and the support for the stinger was lifted. The result of this experiment can be seen in Figure \ref{fig:stinger_frame}, and shows that the robot can support itself through tension.

\subsection{Step response}

\begin{figure}[!h]
    \centering
    \includegraphics[width=0.95\linewidth]{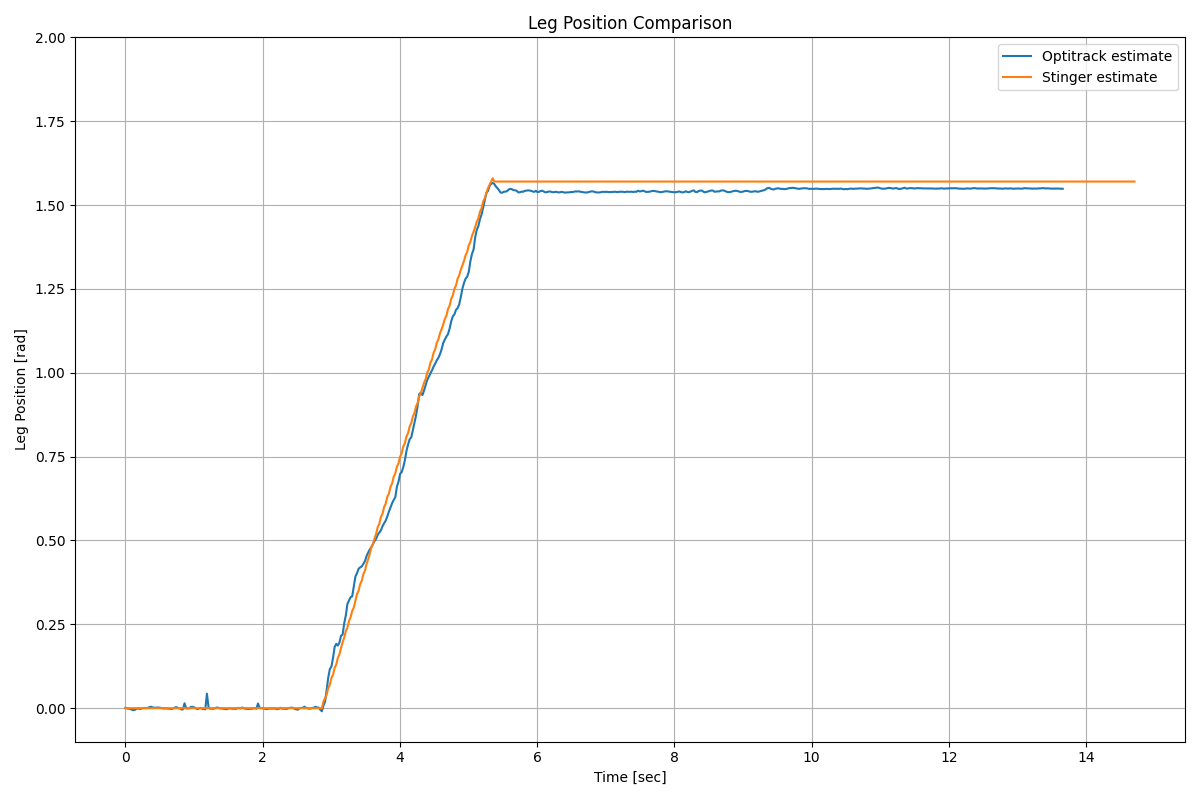}
    \caption{Angular position tracking of one of the Stinger Robot’s rotary joints in response to a 1.57 rad step input compared to Optitrack measurement.}
    \label{fig:step_response}
\end{figure}
The step response test was conducted by using an Optitrack system to achieve a ground truth of the position of one of the stingers of the robot. A 1.57 radians input was then given to the stinger and the response was measured both by the internal encoder on the rotational motor and the Optitrack system. Figure \ref{fig:step_response} shows the results of the step input. It can be seen that the Stinger Robot is able to move its stingers 90 degrees in ~2 seconds and that its rotation estimate is accurate to 0.02 radians.

\section{CONCLUSION}
This paper presented the Stinger Robot, a novel self-bracing robotic platform tailored for autonomous drilling in confined, infrastructure-less underground environments. The system introduces a compact and mechanically robust design that combines a tri-leg, self-locking anchoring mechanism with a force-aware, closed-loop control policy. Through both simulation and preliminary hardware experiments, we demonstrated the robot’s ability to autonomously adapt to irregular geometries, establish stable anchoring, and perform force-intensive tasks such as drilling without external support.
These results validate the feasibility of a new class of compact, adaptive robots capable of operating in environments that are currently inaccessible to traditional mining equipment. Future work will focus on real-world deployment in operational mine settings and full integration into a modular robotic fleet, with the aim of enabling safe, autonomous mineral exploration in previously unreachable locations.

\addtolength{\textheight}{-12cm}   








\end{document}